\documentclass[conference]{IEEEtran}
\usepackage{upgreek}             

\usepackage{ifpdf}

\ifCLASSINFOpdf
   \usepackage[pdftex]{graphicx}
  \graphicspath{{../pdf/}{../jpeg/}}
  \DeclareGraphicsExtensions{.pdf,.jpeg,.png}
\else
  \usepackage[dvips]{graphicx}
 \graphicspath{{../eps/}}
 \DeclareGraphicsExtensions{.eps}
\fi
\usepackage[cmex10]{amsmath}
\begin{document}

%
\title{ Zipf's Law and the  Frequency of Characters or Words of Oracles}

\author{\IEEEauthorblockN{Wang Xiuli}
\IEEEauthorblockA{Anhui University\\
Hefei, Anhui 230039 China\\
Email: wangxiuli@ahu.edu.cn}}
\maketitle

\begin{abstract}
The article discusses the frequency of characters of Oracle,concluding that the frequency and the rank of a word or character is fit to Zipf-Mandelboit Law or Zipf's law with three parameters,and figuring out the parameters based on the frequency,and pointing out that what some researchers of Oracle call the assembling on the two ends is just a description by their impression about the Oracle data.
\end{abstract}

%
\IEEEpeerreviewmaketitle

\bibliographystyle{plain}

\footnotetext[0]{\hspace*{-2em}\small\centerline{\thepage\ --- \pageref{LastPage}}}%

\section{Introduction}

It is known that the American linguist George Kingsley Zipf  presented the law about frequency of word and the rank of word according to it's frequency,the Zipf's law has  experienced much intensive study in different domains by different researchers such as Levy，and Mandelbroit,and hence has been modified to fit into data more precisely,therefore there are varieties of it ,one variety of the law is  $$f=\frac{c}{r^{\alpha}}$$
where $f$ is the frequency,$r$ is the rank,and $\alpha$ is a parameter ,and $c$ is a constant.
another variety is $$f=\frac{c}{(r+a)^{\alpha}} $$ where $a$ is another parameter.

George Kingsley Zipf argues that the law is valid because the least effort principle in human behavior~\cite{Zipf1949}.A recent research shows that Simultaneous minimization in the effort of both hearer and speaker based on the game-theoretic evolution will lead to Zipf's law  in the transition between referentially useless systems and indexical reference systems~\cite{Cancho04022003}.

Chinese characters of Oracles are Chinese archaic characters that are different from modern Chinese characters and inscribed on animal bones and tortoise shell.Since archaic Chinese is single-syllable language that is almost one syllable corresponds to at least one word,and one character represents one word usually .Hence We can know the frequency of words by just counting the frequency of characters.And scholars of Oracle have made some ambiguous or even wrong claim about the distribution of  characters of Oracle such as "assembling on two ends".

In this paper, we will first list some varieties of Zipfs' law,and then give the distribution of words of oracles based on the data collected from transliteration of oracles.We show the distribution is Zipfs' law with three parameter, give the value of the parameters ,and explain why these parameters take such values
\section{Zipfs Law}
Zipfs' Law is an empirical law, there are varieties of it,here we discuss some varieties that are relevant to our study.
\subsection{Zipfs' law with $\alpha=1$}
one variety  of Zipf Law or the original one is the following
$$f=\frac{c}{r^{\alpha}}$$
The following example is list about frequency and rank of  words of Mandarin
 \emph{example 1}, see the figure 2 in the bottom.
\subsection{Zipf law with $\alpha=1$}
If series $$F(\alpha)=\sum_{i=1}^{\infty} f_i=\sum_{i=1}^{\infty} \frac{c}{r_{i}^{\alpha}}$$ \rm is defined on the complex plane,that is $\alpha \in \mathcal{C}$,it is what mathematicians call Riemann $\zeta$ function which has a lot of results to be applicable to areas relevant to Zipf law or the like

there is another variety called Zipf-Mandelbroit formula：
$$f_i =\frac  {c}{(a+r)^{\alpha}}$$
where $0 \leq a < 1$，these can be regarded as terms of the generalized harmonic series:
$$H(\alpha) =\sum_{i=1}^{\infty} f_i=\sum_{i=1}^{\infty} \frac  {c}{(a+r)^{\alpha}}$$
\section{The word frequency of Archaic Chinese and Zipf Law}
\subsection{Word frequency of archaic Chinese of the classical Chinese documents and Zipf Law}
There are some corpus which are  in time near to oracle in history,we have counted the frequency of words of  a work called ShiJi which means it is the record of the history before it's completion ,it was written in about B.C,the time of it's writing is near Shang dynasty relatively in the term of inherited documents . In order to make comparison  with word frequency of oracles,we list the frequency and the rank of words of ShiJi in the following :

 \emph{example 2} word frequency of ShiJi（Part）,see the figures 3-5 in the bottom.

\rm Obviously,Zipf law is valid for ShiJi。

\subsection{ the  Frequency of Characters or Words of Oracles and Zipf's Law }
The language used in oracles is archaic Chinese,but the oracles that scholars had found and collected are in  about several hundred years long time from Shang dynasty(from 1600 BC to 1046 BC) if we do not consider about the oracle bones of  Zhou dynasty.It can be regarded as corpus of archaic Chinese of that period,so we expect that the corpus is fit into Zipf law.

The following is data of \emph{frequency of characters of Oracle } \rm by The Center for Studying and Application of Chinese Character of Huadong Normal University:

Based on the data above,we make them fit to Zipf-Mandelbroit formula  and by caculating,get the following:
$$f_i =\frac  {c}{(a+r)^{\alpha}}$$  $\alpha \approx 1,a \approx 1$
where it's parameters are $\alpha \approx 1,a \approx 1$.We can conclude from Zipf-Mandelbroit formula that is fit to the data above，the corpus of oracle or the language is different from Mandarin or ShiJi,$\alpha \approx 1,a \approx 1$， the parameter reveals that there should be a word with higher frequency than  the word with the highest frequency in the corpus.It seems the word should be a commas,but We know varieties of Zipf law are controversial over the explanation  of it's physics meaning.

\section{conclusion}
word frequency or character frequency of oracle is fit to Zipf-Mandelbroit formula as following:
$$f_i =\frac  {c}{(a+r)^{\alpha}}$$  where $\alpha \approx 1,a \approx 1$.And some researches' conclusion that" the distribution of characters or character frequency of oracle assembles on the two end" is just a description by impression

\centerline{\rule{80mm}{0.1pt}} \vspace{2mm}
\bibliography{reference}

\section*{Appendix}
\begin{figure}[hbt]
\begin{center}
\setlength{\unitlength}{0.0105in}%

\includegraphics[width=2.5in]{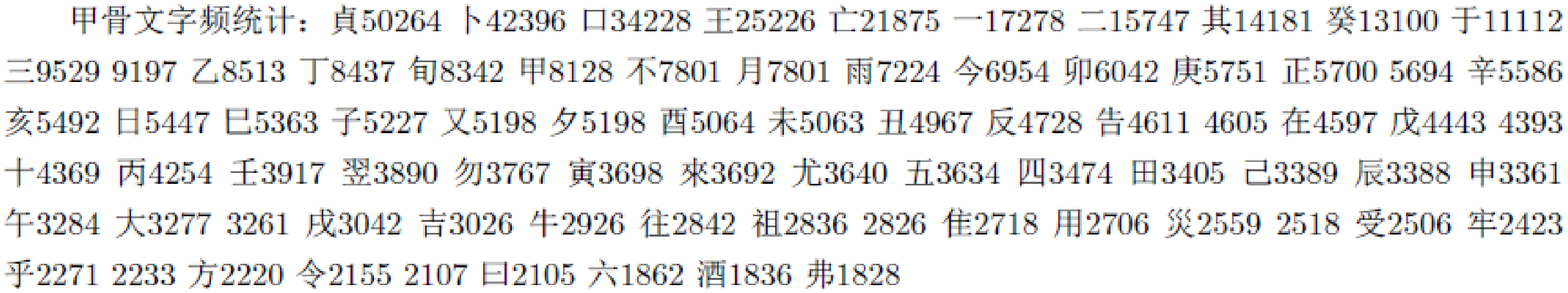}
\end{center}
\caption{The hierarchy of the rational,algebraic,and transcendental number.}
\end{figure}

\begin{figure}[hbt]
\begin{center}
\setlength{\unitlength}{0.0105in}%

\includegraphics[width=2.5in]{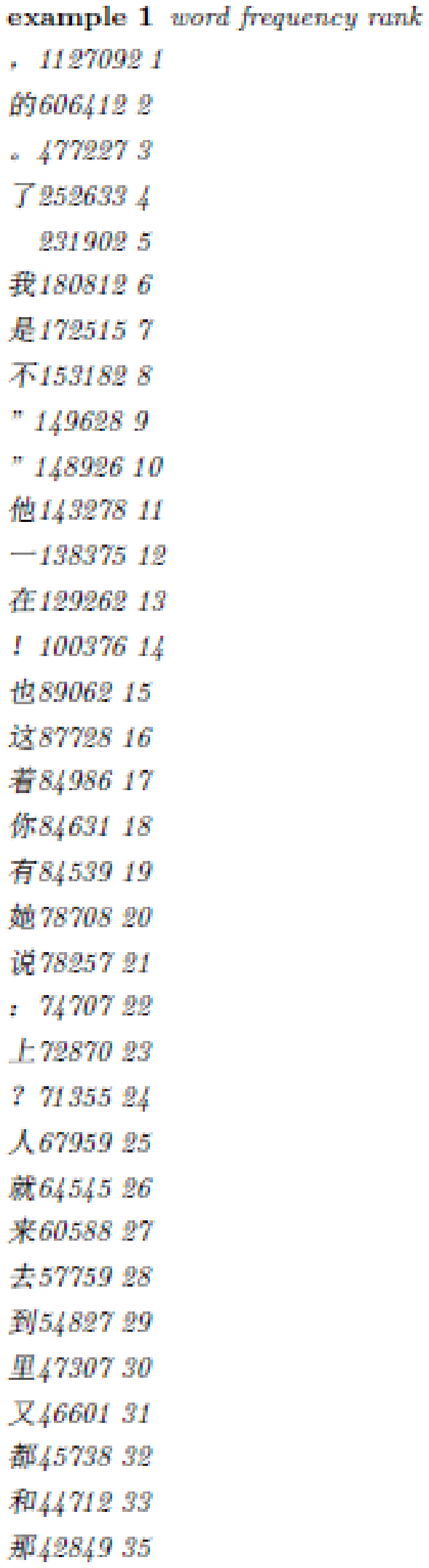}

\end{center}
\caption{Example One.}
\end{figure}

\begin{figure}[hbt]
\begin{center}
\setlength{\unitlength}{0.0105in}%

\includegraphics[width=2.5in]{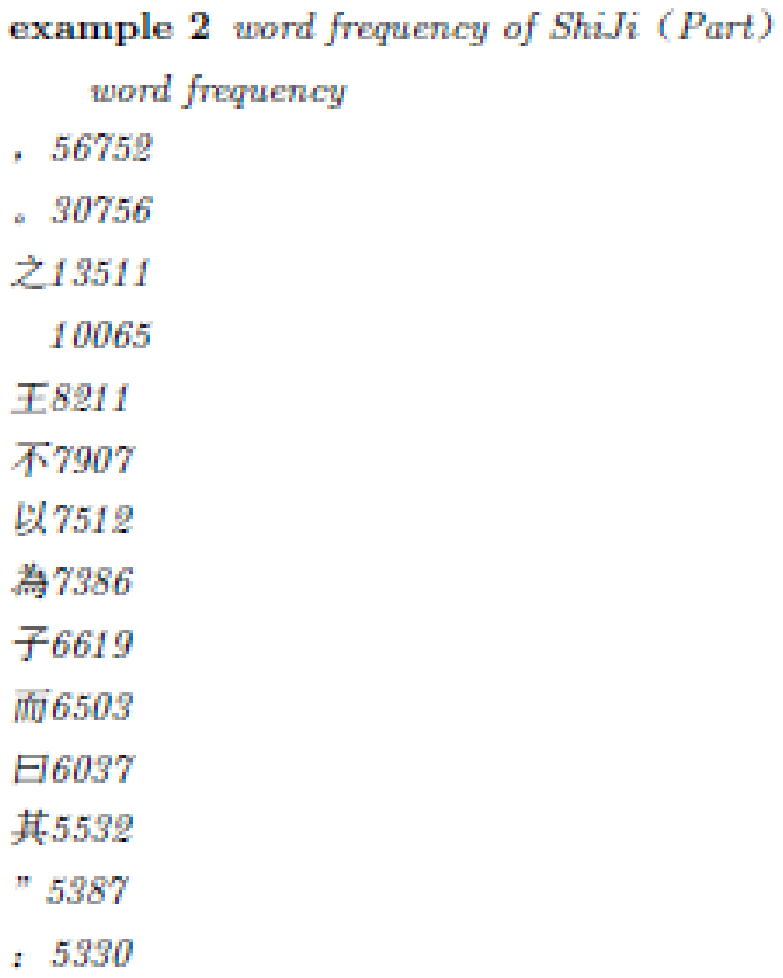}
\end{center}
\caption{ExampleTwo.}
\end{figure}
\begin{figure}[hbt]
\begin{center}
\setlength{\unitlength}{0.0105in}%

\includegraphics[width=2.5in]{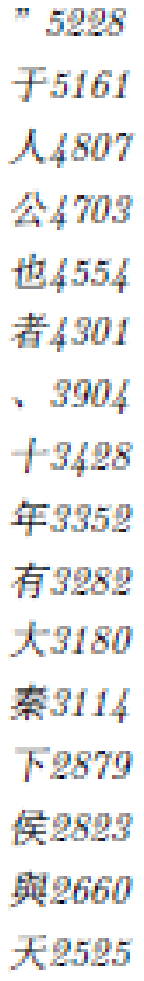}
\end{center}
\caption{ExampleTwo.}
\end{figure}

\begin{figure}[hbt]
\begin{center}
\setlength{\unitlength}{0.0105in}%

\includegraphics[width=2.5in]{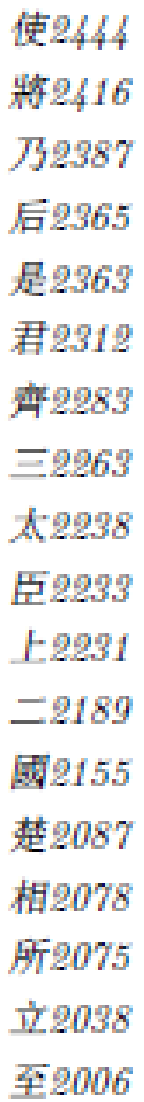}
\end{center}
\caption{ExampleTwo.}
\end{figure}

\footnotetext[0]{Received date Jun. 2014}

\clearpage
\end{document}